\documentclass[sigconf]{acmart}

\usepackage{booktabs} 
\usepackage[linesnumbered,boxed,ruled]{algorithm2e}
\usepackage{graphicx, subfigure}
\usepackage{float}
\usepackage{enumitem}
\usepackage{comment}%
\usepackage{amsmath,amssymb}
\usepackage{hyperref}       

\copyrightyear{2018} 
\acmYear{2018} 
\setcopyright{acmcopyright}
\acmConference[CIKM '18]{The 27th ACM International Conference on Information and Knowledge Management}{October 22--26, 2018}{Torino, Italy}
\acmBooktitle{The 27th ACM International Conference on Information and Knowledge Management (CIKM '18), October 22--26, 2018, Torino, Italy}
\acmPrice{15.00}
\acmDOI{10.1145/3269206.3272010}
\acmISBN{978-1-4503-6014-2/18/10}

\begin{document}
\title{Heterogeneous Graph Neural Networks \\ for Malicious Account Detection}

\author{Ziqi Liu}
\orcid{1234-5678-9012}
\affiliation{%
  \institution{Ant Financial Services Group}
  \city{Hangzhou} 
  \country{China} 
}
\email{ziqiliu@antfin.com}

\author{Chaochao Chen}
\affiliation{%
  \institution{Ant Financial Services Group}
  \city{Hangzhou} 
  \country{China} 
}
\email{chaochao.ccc@antfin.com}

\author{Xinxing Yang}
\affiliation{%
  \institution{Ant Financial Services Group}
  \city{Beijing} 
  \country{China}}
\email{xinxing.yangxx@antfin.com}

\author{Jun Zhou}
\affiliation{%
  \institution{Ant Financial Services Group}
  \city{Beijing}
  \country{China}
}
\email{jun.zhoujun@antfin.com}
\author{Xiaolong Li} 
\affiliation{%
 \institution{Ant Financial Services Group}
 \city{Seattle} 
 \country{USA}}
\email{xl.li@antfin.com}

\author{Le Song}
\affiliation{%
  \institution{Georgia Institute of Technology\\ Ant Financial Services Group}
  \state{GA}
  \country{USA}
}
\email{lsong@cc.gatech.edu}

\renewcommand{\shortauthors}{Ziqi Liu et al.}

\begin{abstract}
We present, GEM, the first heterogeneous graph neural network approach for detecting
malicious accounts at Alipay, one of the world's leading mobile cashless payment platform.
Our approach, inspired from a connected subgraph approach, adaptively learns discriminative
embeddings from heterogeneous account-device
graphs based on two fundamental weaknesses of attackers, i.e. device aggregation and
activity aggregation. For the heterogeneous graph consists of various types of nodes,
we propose an attention mechanism to learn the importance of different types of nodes,
while using the sum operator for modeling the aggregation patterns of nodes in each type.
Experiments show that our approaches consistently perform
promising results compared with competitive methods over time.
\end{abstract}

%
%
%

\ccsdesc[500]{Computing methodologies~Neural networks}

\keywords{Malicious account detection; neural graph networks; heterogeneous graphs}

\maketitle

\section{Introduction}\label{sec:intro}
Large scale online services such as Gmail\footnote{https://mail.google.com},
Facebook\footnote{https://www.facebook.com} and Alipay\footnote{http://render.alipay.com/p/s/download}
have becoming popular targets for cyber attacks. By creating malicious accounts,
attackers can propagate spam messages, seek excessive
profits, which are essentially harmful to the eco-systems.
For example, numerous abused bot-accounts were used to send out billions of spam
emails across the email system. What is more serious is that in financial systems like
Alipay, once a large number of accounts be taken over by a malicious user or a group of them,
those malicious users could possibly cash out and gain ill-gotten earnings, that
enormously harms the whole financial system. Effectively and accurately detecting such malicious accounts
plays an important role in such systems.

Many existing security mechanisms to deal with malicious accounts
have extensively studied the attack characteristics
~\cite{xie2008spamming, zhao2009botgraph, huang2013socialwatch, cao2014uncovering, stringhini2015evilcohort} which
hopefully can discern the normal and malicious accounts.
To exploit such characteristics, existing research mainly spreads in three directions.
First, \emph{Rule-based methods} directly generate sophisticated rules for identification.
For example, Xie et al.~\cite{xie2008spamming} proposed ``spam payload'' and ``spam server traffic''
properties for generating high quality regular expression signatures.
Second, \emph{Graph-based methods} reformulate the problem by considering the connectivities
among accounts. This is based on the intuition that attackers can only evade individually
but cannot control the interactions with normal accounts.
For example, Zhao et al.~\cite{zhao2009botgraph} analyzed connected subgraph components by constructing
account-account graphs to identify large abnormal groups.
Thrid, \emph{Machine learning-based methods} learn statistic models by exploiting large amount
of historical data. For examples,
Huang et al.~\cite{huang2013socialwatch} extracted features based on graph properties
and built supervised classifiers for identifying malicious account.
Cao et al.~\cite{cao2014uncovering} advanced the usages of aggregating behavioral
patterns to uncover malicious accounts in an unsupervised machine learning framework.

As attacking strategies from potential adversaries change, it is crucial that
a well-behaved system could adapt to the evolving strategies~\cite{zhao2009botgraph, cao2014uncovering}.
We summarize the following two major observations from attackers as
the fundamental basis of our work.
(1) \emph{Device aggregation}. Attackers are subjected to cost on computing resources. That is,
due to economic constraints, it is costly if attackers can control
a large amount of computing resources. As a result, most accounts
owned by one attacker or a group of attackers will signup or sigin frequently on only a small number
of resources. 
(2) \emph{Activity aggregation}. Attackers are subject to the limited time of campaigns. Basically,
attackers are required to fulfil specific goals in a short term.
That means the behaviors of malicious accounts controlled
by a single attacker could burst in limited
time.

The weaknesses of attackers have been extensively analyzed, however, 
it's still challenging to identify attackers with both high
precision and recall\footnote{https://en.wikipedia.org/wiki/Precision\_and\_recall}.
In financial systems like Alipay, it is way important to \emph{accurately} identify
malicious account \emph{as many as possible}. The reason is in two-folds: (1) The illegal
behaviors like cash-out is essentially harmful to the whole financial system or even
the national security; (2) As an Internet service
company, we need to reduce the unnecessary disturbances and interruptions to normal users, i.e. providing friendly services.
Existing methods~\cite{zhao2009botgraph} usually achieve very low \emph{false positive rate} (friendly services)
by setting strict constraints but potentially missing out the opportunities on identifying
much more suspicious accounts, i.e. with a \emph{high false negative rate}.
The reason is that the huge amount of benign accounts interwined with
only a small number of suspicious accounts, and this results into a low
signal-to-noise-ratio. It is quite common that normal accounts share the same
IP address with malicious accounts due to the noisy data, or the IP
address comes from a common proxy. Thus make it important to
jointly consider the ``Device aggregation'' and ``Activity
aggregation'' altogether in the view of \emph{heterogeneous} graph consists of various
types of devices such as phone number, Media access control address (MAC),
IMEI (International Mobile Equipment Identity), SIM number, and so on.

In this work, we present, \textbf{G}raph \textbf{E}mbeddings for \textbf{M}alicious accounts (GEM),
a novel nueral network-based graph technique based on the literature of graph representation
learning~\cite{hamilton2017representation}, which \emph{jointly} considers
``Device aggregation'' and ``Activity aggregation''
in heterogeneous graphs. Our proposed approach essentially models the topology of 
the heterogeneous account-device graph, and simultaneously considers the characteristics of
activities of the accounts in the local strucuture of this graph. The basic idea
of our model is that whether a single account is normal or malicious is a function
of how the other accounts ``congregate'' with this account via devices
in the topology, and how those other accounts shared the same
device with this account ``behave'' in timeseries. To allow various types of devices,
we use attention mechanism to adaptively learn the importance of different types of devices.
Unlike existing methods that one first studies the graph properties~\cite{huang2013socialwatch}
or pairwise comparisons of account activities~\cite{cao2014uncovering}, then feeds into a
machine learning framework, our proposed method directly learns a function for each
account given the context of the local topology and other accounts' activities nearby
in an end to end way.

We deploy the proposed work as a real system at Alipay. It can detect tens of thousands
malicious accounts daily. We empirically show that the experimental results
significantly outperform the results from other competitive methods.

We summarize the contributions of this work as follows:
\begin{itemize}
\item We present a novel neural network based graph representation method for identifying
  malicious accounts by jointly capturing two of attackers' weaknesses, summarized as
  ``Device aggregation'' and ``Activity aggregation'' in a heterogeneous graph.
  To our best knowledge, this is the first fraud detectoin problem addressed by
graph neural network approaches with careful graph constructon.

\item Our approach is deployed at Alipay, one of the largest
  third-party mobile and online cashless payment platform serving
  more than 4 hundreds of million users. The approach can detect
  tens of thousands malicious accounts daily.
\end{itemize}

\section{Preliminaries}
In this section, we first briefly present some preliminary contents
of graph representation learning techniques recently developed.

\subsection{Graph Neural Networks}
The first class is concerned with predicting labels over a graph,
its edges, or its nodes. Graph Neural Networks were introduced
in Gori et al.\cite{gori2005new} and Scarselli et al. \cite{scarselli2009graph} as a generalization of recursive
neural networks that can directly deal with a more general class of graphs,
e.g. cyclic, directed and undirected graphs.

Recently, generalizing convolutions to graphs have shown promising results~\cite{bruna2013spectral,defferrard2016convolutional}.
For example, Kipf \& Welling \cite{kipf2016semi} propose simple filters that operate in a
1-step neighborhood around each node. Assuming $X \in \mathbb{R}^{N,D}$ is a matrix of
node features vectors $x_i \in \mathbb{R}^D$, an undirected graph $\mathcal{G}=(\mathcal{V}, \mathcal{E})$
with $N$ nodes $v_i \in \mathcal{V}$, edges $(v_i, v_j) \in \mathcal{E}$,
an adjacency matrix $A \in \mathbb{R}^{N\times N}$. They propose the following
convolution layer:
\begin{align}\label{eq:convolution}
  H^{(t+1)} = \sigma\bigg( \tilde{A} H^{(t)} W^{(t)} \bigg), 
\end{align}
where $\tilde{A}$ is a symmetric normalization of $A$ with self-loops, i.e.
$\tilde{A}=\hat{D}^{-\frac{1}{2}}\hat{A}\hat{D}^{-\frac{1}{2}}$,
$\hat{A}=A+I$ and $\hat{D}$ is the diagonal node degree matrix of $\hat{A}$, $H^{(t)}$ denotes
the $t$-th hidden layer with $H^{(0)}=X$, $W^{(t)}$ is the layer-specific
parameters, and $\sigma$ denotes the activation functions.
The GCN~\cite{kipf2016semi} essentially learn a function $f(X,A)$ that helps the
representation of each node $x_i$ by exploiting its neighborhood defined in $A$.
By modeling nodes as documents, and edges as citation links, their algorithm achieves state-of-art
results on tasks of classifying documents in citation networks like Citeseer, Cora,
and Pubmed.

At the same time, a novel connection between graphical models and neural networks
has been studied by Dai et al.~\cite{dai2016discriminative}. One key observation
is that the solution of variational latent distribution $q_i(\mu_i)$ for each node $i$  
needs to satisfy the following fixed point
equations:
\begin{align}
  \log q_i(\mu_i) = g(\mu_i, x_i, \{q_j\}_{j\in \mathcal{N}(i)}).
\end{align}
Moreover, Smola et al.~\cite{smola2007hilbert} showed that there
exists another feature space such that one can find an injective embedding
$h_i$ as sufficient statistics corresponding to the original $q_i(\mu_i)$ function.
As a result, Dai et al.~\cite{dai2016discriminative} shows that for any given above
fixed point equation one can always find an equivalent
transformation in another feature space:
\begin{align}
  h_i = f(x_i, \{h_j\}_{j \in \mathcal{N}(i)}).
\end{align}
As such, one can directly learn the graphical model in
the embedding space $h$ and directly optimize
the funtion $f$ by extra link functions in
a neural network framework. Such representation is even more
powerful compared with traditional graphical models where
each variable is limited by a function from
an exponential family.

\begin{figure*}
\centering
\subfigure { \includegraphics[width=0.45\textwidth, height=0.31\textwidth]{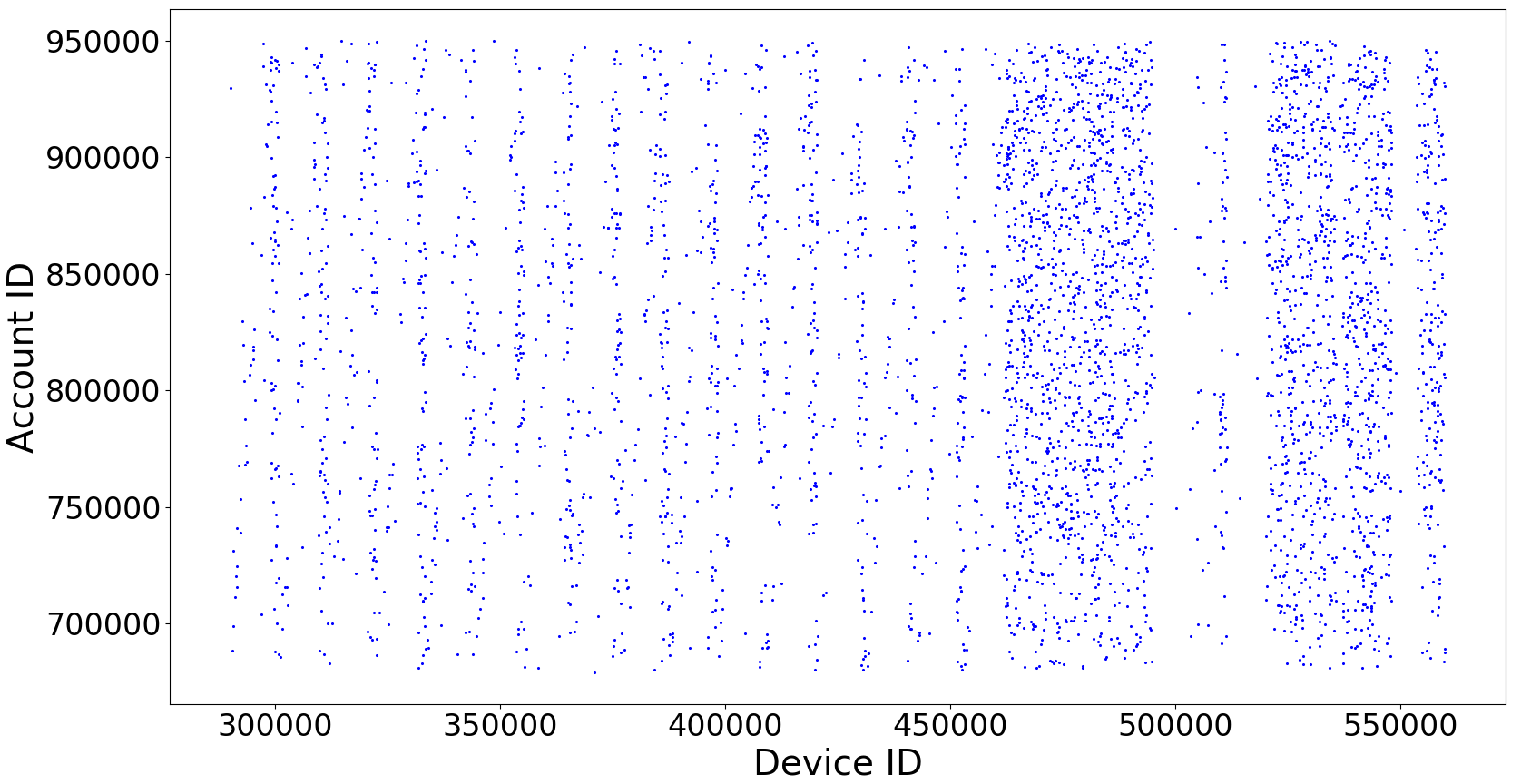}}
\subfigure { \includegraphics[width=0.45\textwidth, height=0.31\textwidth]{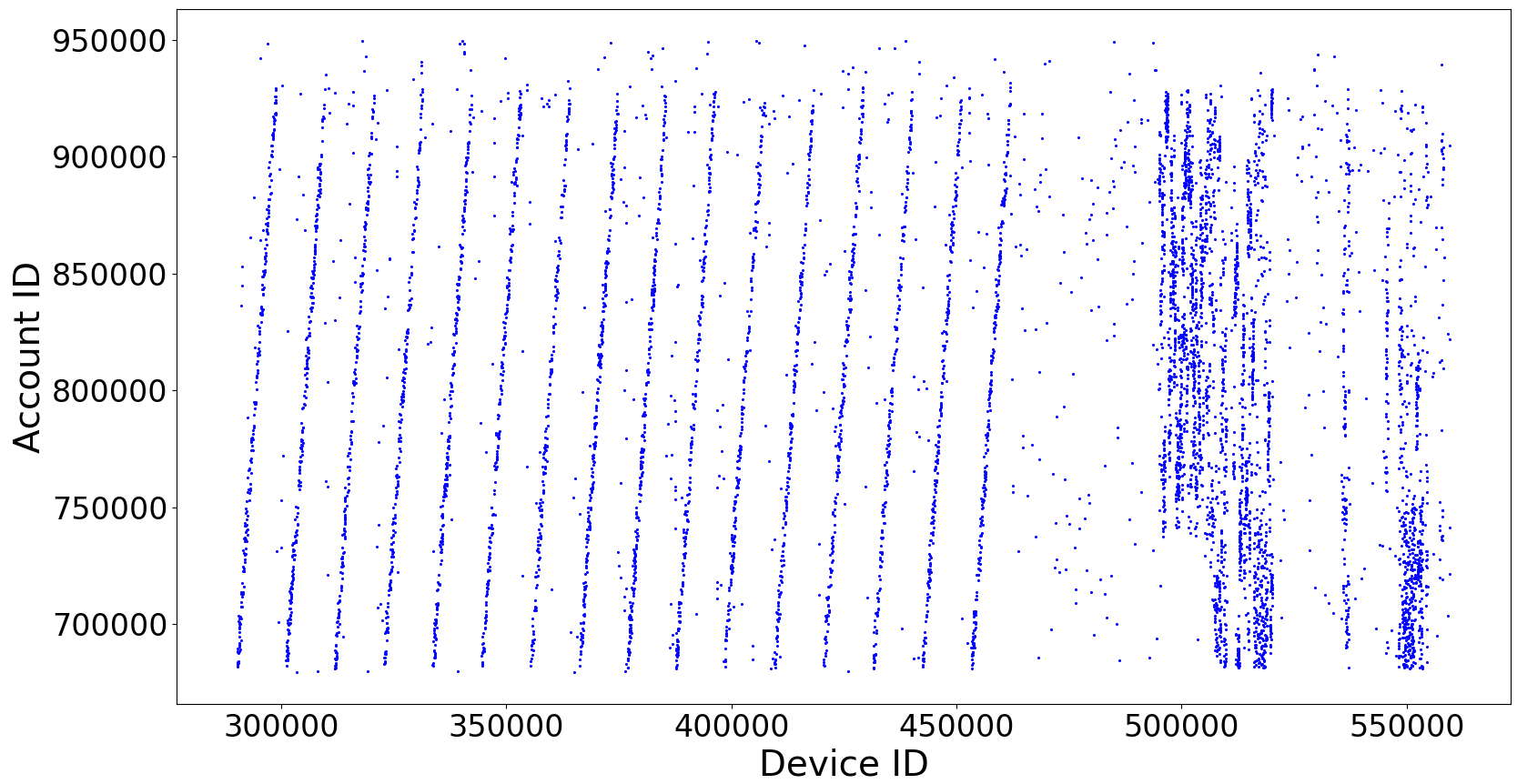}}
\caption{Device aggregation patterns: normal accounts (\emph{first}) v.s.
  malicious accounts (\emph{second}), malicious accounts tend to aggregate compared with normal accounts.}
\label{fig:pattern1}
\end{figure*}

To summarize, the works in this domain essentially are built based on
an iterative-style neighborhood aggregation method~\cite{hamilton2017representation}:
\begin{align}
  h_i^{(t+1)} = f(x_i, x_{j \in \mathcal{N}(i)}, h_i^{(t)}, h_{j \in \mathcal{N}(i)}^{(t)})
\end{align}
where $f(\cdot)$ is a parameterized non-linear transformation.
Most of the efforts in this domain study the ``receptive fields''~\cite{liu2018geniepath} 
that aggregation operators should work on, because compared with data like images
where each pixel have exactly 8 neighbors, the nodes in the
graph domain can vary a lot.

More recently, Liu et al.~\cite{liu2018geniepath} propose GeniePath, which
aims to adaptively filter each node's receptive fields, compared with
GCN that does convolution on pre-defined receptive fields. And this yields
much better results.

Our methods can be viewed as a
variant of graph convolutional networks,
that is, we design a approach that uses the sum operator to
capture the ``aggregation'' patterns in each node's $T$-step neighborhood,
while using attention mechnisms to reweigh the importances of
various types of nodes in the ``heterogeneous'' graph.

\subsection{Node Embedding}\label{sec:node2vec}
The second class of techniques consists of graph embedding methods
that aim to learn representation of each node while preserving the graph
structure~\cite{hamilton2017representation}.
They explicitly model the relationships among node pairs.
For example, some methods directly use the adjacency
matrix~\cite{ahmed2013distributed, belkin2002laplacian},
$t$-th order adjacency matrix~\cite{cao2015grarep}, 
and others simulate random walks by approximating the
high order adjacency matrix in a randomized
manner~\cite{grover2016node2vec, perozzi2014deepwalk}.

Formally, most approaches aim to minimize an emprical loss,
$\mathcal{L}$, over a set of training node pairs:
\begin{align}
  \mathcal{L} = \sum_{(v_i, v_j) \in \mathcal{E}}
  \ell(f_{\mathrm{dec}}(f_{\mathrm{enc}}(v_i), f_{\mathrm{enc}}(v_j)), f_{\mathcal{G}}(v_i, v_j)),
\end{align}
where $f_{\mathrm{enc}}: \mathcal{V} \mapsto \mathbb{R}^k$ is the encoder function,
$f_{\mathrm{dec}}: \mathcal{V} \times \mathcal{V} \mapsto \mathbb{R}^+$ is the decoder function such as
$f_{\mathrm{dec}}(z_i, z_j) = z_i^\top z_j$, and $f_{\mathcal{G}}(\cdot, \cdot) \mapsto \mathbb{R}^+$ is
the so called pairwise proximity function, and $\ell: \mathbb{R} \times
\mathbb{R} \mapsto \mathbb{R}$ is a specific loss function used to measure
the reconstruction ability of $f_{\mathrm{dec}}, f_{\mathrm{enc}}$ to a user-specified pairwise
proximity measure $f_{\mathcal{G}}$.

The methods in this domain are unsupervised algorithms. They learn
node embeddings on the graph without use of ground truth labels.
Such node embeddings can be used as statistic properties of the graph
as~\cite{huang2013socialwatch}, and be fed into a classifier for
final prediction.

Pratically the random walk-based proximity
measure~\cite{hamilton2017representation} has
proven to achieve state-of-art
results on many tasks like citation networks, protein networks etc.
We will report the empirical results of such methods in experiments.

\begin{figure*}
\centering
\subfigure { \includegraphics[width=0.45\textwidth, height=0.31\textwidth]{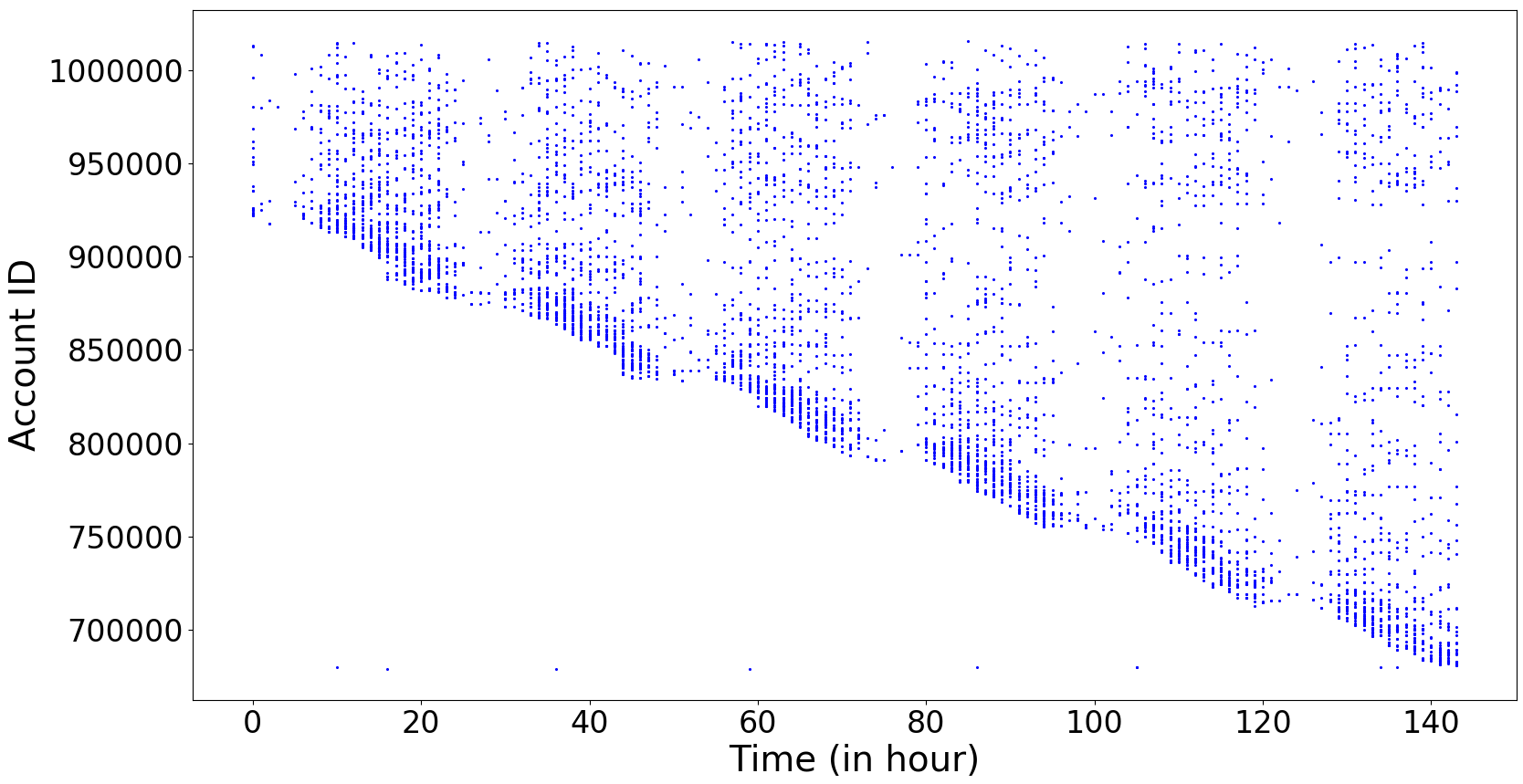}}
\subfigure { \includegraphics[width=0.45\textwidth, height=0.31\textwidth]{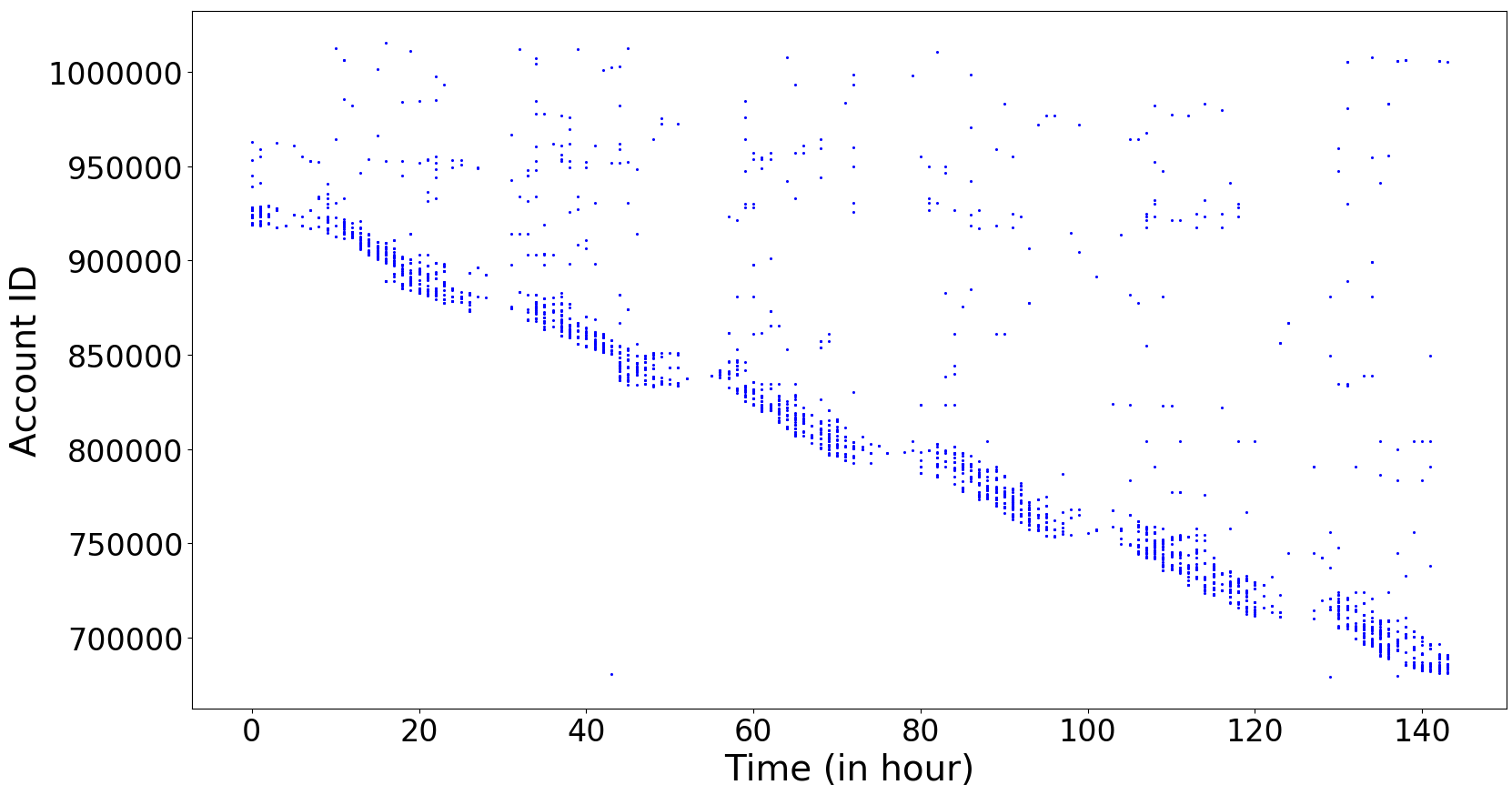}}
\caption{Activity aggregation patterns:
  normal accounts (\emph{first}) v.s. malicious accounts (\emph{second}), newly registered malicious accounts
  tend to be active only in a short term.}
\label{fig:pattern2}
\end{figure*}

\begin{figure}
\centering
\subfigure { \includegraphics[width=0.4\textwidth, height=0.51\textwidth]{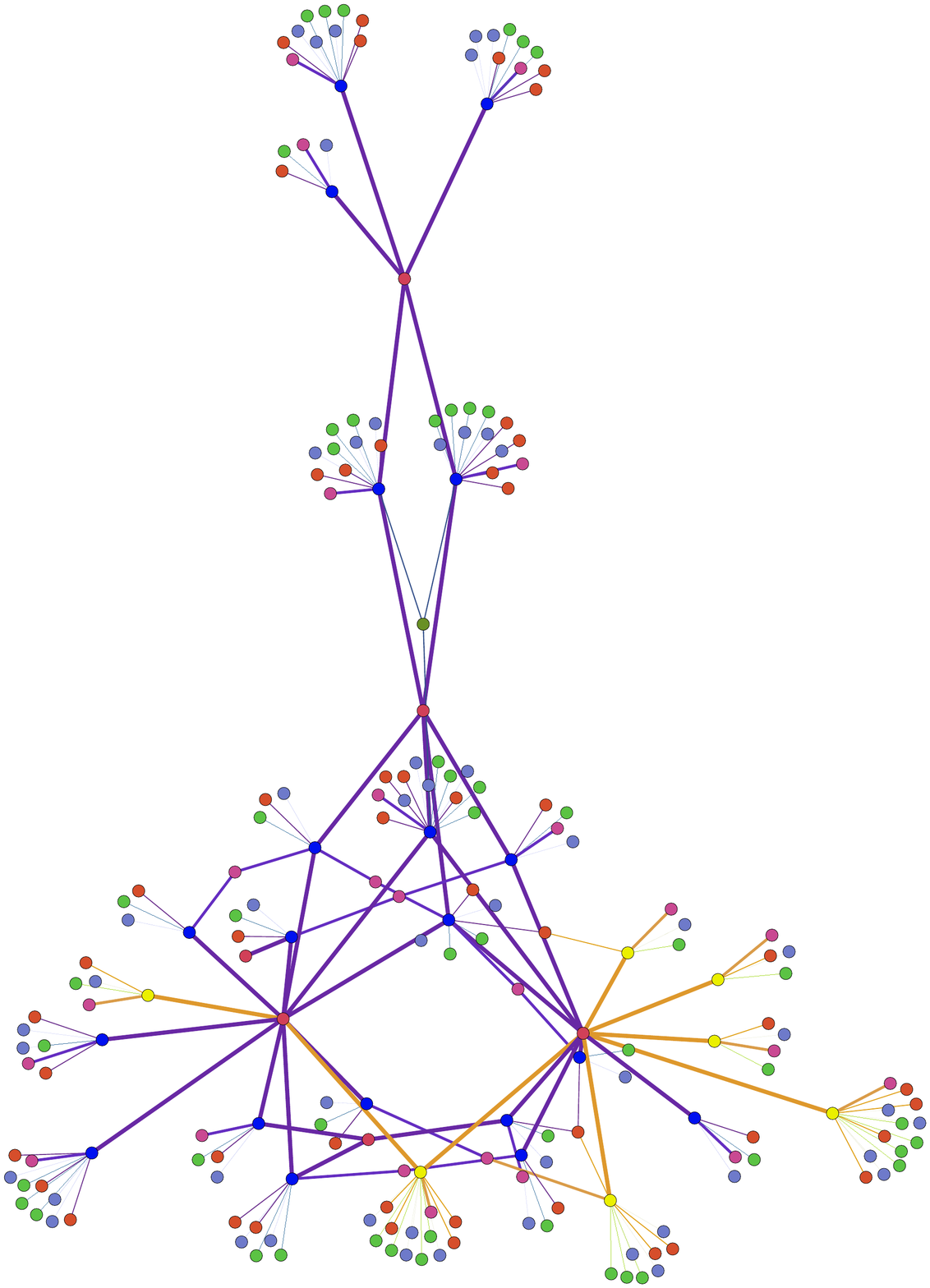}}
\caption{Visualization of one of connected subgraph component from our data, a total of 210 vertices consists of 20 normal accounts (\emph{Blue}),
7 malicious accounts (\emph{Yellow}), and rest of vertices correspond to 6 types of devices. The thicknesses of those edges
denote the estimated attention coefficients will be discussed in section~\ref{sec:attention}.}
\label{fig:visual}
\end{figure}
\section{The Proposed Approaches}
In this section, we first describe the patterns we found in
the real data at Alipay, then discuss a motivated approach based on
connected subgraph components. Inspired from this intuitive approach, we discuss the construction of
a heterogeneous graph based on the characteristics of the real data, and finally
present the approach on modeling malicious accounts.

\subsection{Data Analysis}\label{sec:example}
In this section we study the patterns of ``Device aggregation'' and ``Activity aggregation''
demonstrated by the real data at Alipay.

\emph{Device aggregation}. The basic idea of device aggregation is that
if an account signups or logins the same device or a common set of devices
together with a large number of other accounts, then such accounts would be suspicious.
One can simply calculate the size of the connnected subgraph components~\cite{zhao2009botgraph}
as a measure of risk for each account.

\emph{Activity aggregation}. The basic idea of activity aggregation is
that if accounts sharing the common devices behave in batches, then those
accounts are suspicious. One can simply define the inner product
of activities of two accounts sharing the same device as a measure of affinity, i.e.
$S_{i,i'}^a = \langle x_i, x_{i'}\rangle$. Apparently the consistent
behaviors over time between account $i$ and $i'$ mean high degrees
of affinity.
Such measures of affinity between two accounts can be further used
to split a giant connected subgraph to improve the false positive rate~\cite{zhao2009botgraph}.

We illustrate such two patterns from
the data of Alipay in Figure~\ref{fig:pattern1} and Figure~\ref{fig:pattern2}.
Figure~\ref{fig:pattern1} shows account-device graphs accumulated
in 7 consecutive days. We do not differentiate the different types of device in this graph.
A blue dot means the account has behaviors (signups or logins)
associated with the corresponding device. For normal accounts, the blue dots
uniformly scatter over the account-device graph, compared with malicious accounts,
the blue dots show strong signals that the specific device could connect with huge number
of accounts in various patterns. Figure~\ref{fig:pattern2} illustrates the behavior patterns of each account over time,
where each blue dot denotes that there is an activity of account $i$ at time $t$.
The behaviors of normal accounts in graph on the left show that each newly registered normal
accounts behave evenly in the next several days, whereas the malicious accounts in
the second graph show that they tend to burst only in a short time.

The patterns analyzed in this section motivate us the consideration of
modeling malicious accounts in the view of graph.


\subsection{A Motivation: Subgraph Components}\label{sec:mot}
The device aggregation pattern in Figure~\ref{fig:pattern1} and activity aggregation
pattern in Figure~\ref{fig:pattern2} inspired us the consideration of the problem in graph.

We call our first attempt as ``Connected Subgraph''. Our basic idea is to build a graph
of accounts, hopefully with edges connect a gang of accounts. The ``Connected Subgraph'' approach consists of three steps:
\begin{enumerate}
\item Assume we have a graph $\mathcal{G}=(\mathcal{V}, \mathcal{E})$, with $N$ nodes include accounts
  and devices, and a set of $M$ edges $\{(i,j)\}$ denote login behaviors of account $i$ on device $j$
  during a time period. We aim to build a homogeneous graph $\mathcal{G}^a = (\mathcal{V}^a, \mathcal{E}^a)$,
  consists of only accounts as nodes. That is, we add an edge $(i,i')$ if there exists $(i,j)$ and $(i',j)$
  that both account $i$ and $i'$ login the same device during a time period. As such, the homogeneous graph
  $\mathcal{G}^a$ consists of connected subgraphs with each subgraph somehow captures a group of accounts.
  The larger the group, the potentially higher risk this group could be a gang of malicious accounts.
  However, the data are naturally noisy in practice, and it is quite common that different accounts
  login the same IP addresses and so on, thus interwines normal accounts and malicious accounts.
\item We further reduce and delete the edges as follows. As we see from Figure.~\ref{fig:pattern2},
  the activities of a gang of accounts mostly burst in a short period of a certain day. To measure the
  similarity between two accounts in a subgraph of $\mathcal{G}^a$, we characterize each account $i$'s
  behavior as a vector $x_i = [x_{i,1}, ..., x_{i,p}]^\top$, with $p=24$ and each $x_{i,t}$ denote the
  frequency of behaviors at the $t$-th hour. We could measure the similarity between two accounts as
  the inner product $x_i^\top x_{i'}$. As such, we further delete edges $(i,i')$ of graph $\mathcal{G}^a$ in case
  $x_i^\top x_{i'} < \theta$ where $\theta$ is a hyperparameter controls the sparsity of graph $\mathcal{G}^a$.
\item Finally, we can score each account using the size of the subgraph it belongs to. To determine
  the hyperparameter $\theta$, We can tune $\theta$ on a validation set.
\end{enumerate}

Even though this approach is intuitive and it can accurately detect malicious accounts in the largest
connected subgraphs, its performance deteriorates seriously for those malicious accounts lie in 
smaller subgraphs. 

Is there any way of discerning malicious accounts from normal accounts with a more machine learning oriented
approach? Different from traditional machine learning approach that people first extract useful features $X$,
then learn a discriminate function $f(X)$ to discern those accounts, can we directly learn a function $f(X,\mathcal{G})$
that jointly utilizes the topology of the graph and features?

One observation is that the three steps involved in ``Connected Subgraph'' essentially pre-define a score function $g(\cdot)$ on
each node based on (1) the ``connectivities'' around its neighborhood, and (2) the sum operator that counts the nodes
lie in the connectivities. The connectivities depend both on the topology of graph $\mathcal{G}^a$ (device aggregation), and 
the inner product among nodes (activity aggregation) that further constrains the connectivities. The sum operator measures the
aggregated strength of the connectivities, i.e. the size of the subgraph. Another observation is that
we actually have a function to transform
the orginal account-device graph $\mathcal{G}$ to account-account graph $\mathcal{G}^a$. This step is important
for ``Connected Subgraph'' because else we have no way to measure the affinity among different accounts, however,
the transformation essentially discards informations from the original graph.

In the following sections, we would like to learn a parameterized score function based on the existing graph
representation learning literature. In particular, we are interested in embedding each node into vector spaces,
so as to imitate the sum of ``connectivities'' in the space of $\mathcal{G}^a$.

\subsection{Heterogeneous Graph Construction}\label{sec:het}

Assuming $N$ vertices include accounts and devices with each 
device corresponds to one type $d \in \mathcal{D}$.
We observe a set of $M$ edges $\{(i,j)\}$ among accounts
and devices over a time period $[0,T)$. Each edge denotes that the account
  $i$ has activities, e.g. signup, login and so on, on device $j$. As such,
  we have a graph $\mathcal{G} = (\mathcal{V}, \mathcal{E})$ consists
  of $N$ accounts and devices, with edges connecting them. In terms of
  linear algebra, this leads to an adjacency matrix
  $A \in \{0,1\}^{N, N}$.
  We illustrate one of the connected subgraph of $\mathcal{G}$ from our dataset
  in Figure~\ref{fig:visual}.

  For our convenience, we can further extract
  $|\mathcal{D}|$ subgraphs $\{\mathcal{G}^{(d)} = (\mathcal{V}, \mathcal{E}^{(d)})\}$
  each of which preserves all the vertices of $\mathcal{G}$,
  but ignores the edges containing devices that do not belong to type $d$.
  This leads to $|\mathcal{D}|$ adjacency matrices $\{A^{(d)}\}$.
  Note that the heterogeneous graph representation $\{\mathcal{G}^{(d)}\}$ lies in the same
  storage complexity compared with original $\mathcal{G}$ because we
  only need to store the sparse edges.

Note that the ``device'' here could be a much broader concept. For example,
the device could be an IP address, a phone number,
or even a like page in facebook.
In our data, we collect various types of device include phone number,
User Machine ID (UMID)\footnote{The fingerprint built by Alibaba for uniquely identifying devices.}, MAC address,
IMSI (International Mobile Subscriber Identity), APDID (Alipay Device ID)\footnote{
The fingerprint built by Alipay for uniquely identifying device by considering
IMEI, IMSI, CPU, Bluetooth ADDR, ROM together.}
and TID\footnote{A random number generated via IMSI and IMEI (International Mobile Equipment Identity)}, thus results into a heterogeneous graph.
Such heterogeneous graphs
allow us to understand different implications of different devices.

Along with these graphs, we can further observe the activities
of each account. Assuming a $N$ by $p+|\mathcal{D}|$ matrix
$X \in \mathbb{R}^{N, p+|\mathcal{D}|}$,
with each row $x_i$ denotes activities of vertex $i$ if $i$ is an account. In practice,
the activities of account $i$ over a time period $[0,T)$  can be discretized
into $p$ time slots, where the value of each time slot denotes
the count of the activities in this time slot. For vertices
correspond to devices, we just encode $x_i$ as one hot vector using
the last $|\mathcal{D}|$ coordinates.


Our goal is to discriminate between malicous and normal accounts.
That is, given the adjacency matrix $A$ and activities $X$ during
time $[0,T)$, and $N_o$ partially observed truth labels
$\{y_i\} \in \{-1,1\}$ of accounts only over time $[0,T-1)$, we
aim to learn a function $f(\{A^{d}\}, X)$ to correctly identify malicious
accounts and generalize well on data at time $T$.

\subsection{Models}
In the above sections, we discussed the patterns that we observed in real data,
and the construction of heterogeneous graphs include accounts and
various types of devices. We claimed that ``Device aggregation'' and
``Activity aggregation'' can be learned as a function of the adjacency
matrix $A$ and activities $X$. It remains to specify a powerful
representation of the function to capture those patterns.

In our problem, we hope to learn effective embeddings $h_i$ for
each vertex $i$ by propagating transformed activities $X$
on topology of graphs $\{\mathcal{G}^{(d)}=(\mathcal{V}, \mathcal{E}^{(d)})\}$:
\begin{align}\label{eq:estep}
  & H^{(0)} \leftarrow \bf{0} \\\nonumber
  & \text{for} \,\,\,\,\, t = 1,...,T \\\nonumber
  & \,\,\,\,\,\,\,\,\,\,\,\,H^{(t)} \leftarrow \sigma \bigg( X\cdot W + \frac{1}{|\mathcal{D}|} \sum_{d=1}^{|\mathcal{D}|} A^{(d)} \cdot H^{(t-1)}\cdot V_d \bigg)
\end{align}
where $H^{(t)} \in \mathbb{R}^{N,k}$ denotes the embedding matrix at $t$-th
layer with the $i$-th row corresponds to $h_i$ of vertex $i$, $\sigma$ denotes a nonlinear activation,
e.g. a rectifier linear unit activation function, $W \in \mathbb{R}^{P \times k}$ with $P=p+|\mathcal{D}|$
and $\{V_d\} \in \mathbb{R}^{k \times k}$ are parameters to
control the ``shape'' of the desired function
given the connectivities and related activities of accounts, with the hope that they can automatically
capture more effective patterns. 
We let $k$ denote the embedding size, and $T$ denote the number of hops a vertex needs to look at, or
the number of hidden layers.
As the layers being deeper, i.e. $T$ being larger, for example, $h_i^{(5)} \in \mathbb{R}^k$ means aggregation of
$i$'s neighbors up to 5 hops away.
We allow $x_i$ appears in each hidden layer as per Eq.~\eqref{eq:estep}, that can somehow connect
deep distant layers like in the residual networks~\cite{he2016identity}.
Empirically we set $k=16$ and $T=5$ in our experiments.
We normalize the impact of different types of devices
by averaging, i.e. $\frac{1}{|\mathcal{D}|}$.

{\bfseries Some explainations}. In case we ignore the type of devices $d$ and extent of neighborhood $T$,
the transformation $H V$ in Eq.~\eqref{eq:estep} embeds each account $i$'s activities $x_i$
into a latent vector space, then the operation $A\,HV$ sums the $1$-step neighborhood's latent vectors.
As we iterate this layer after $T$ steps, the operator essentially sum over each node's $T$-step neighborhood
in latent vector spaces, which is similar to the function $g(\cdot)$ defined in ``Connected Subgraph'', that
sums the number of nodes lie in the reachable ``connectivities''.
The difference is that, our approach works on the original account-device graph, and embeds
each node into a latent vector space by summing over its $T$-step neighbors' embedded activities
along the topology. As a result, we can learn a parameterized function governed by only $W$
and $\{V_1,...,V_{|\mathcal{D}|}\}$ in a more machine learning oriented manner.
Without adjacency matrix $A$, our model degenerates to a deep neural network with ``skip connection''
architecture~\cite{he2016identity} that relies only on features $X$.

{\bfseries Optimization}. To effectively learn $W$ and $\{V_1,...,V_{|\mathcal{D}|}\}$, we link those embeddings to a standard logistic
loss function:
\begin{align}\label{eq:mstep}
  \min_{W,\{V_d\},u} \mathcal{L}(W,\{V_d\},u) = -\sum_{i}^{N_o} \log \sigma \big(y_i \cdot (u^\top h_i)\big)
\end{align}
where $\sigma$ denotes logistic function $\sigma(x) = \frac{1}{1+\exp{-x}}$, $u \in \mathbb{R}^{k}$,
and the loss $\mathcal{L}$ sums over partially observed $N_o$ accounts with known labels. 
Our algorithm works interatively in an Expectation Maximization style.
In e-step, we compute the embeddings based on current parameters
$W,\{V_d\}$ as in Eq~\eqref{eq:estep}. In m-step, we optimize those parameters in
Eq~\eqref{eq:mstep} while fixing embedings.

Our approach can be viewed as a variant of graph convolutional networks~\cite{kipf2016semi}.
However, the major difference lies in (1) we generalize the algorithms
to heterogeneous graphs; (2) the aggregation operator defined on the neighborhood.
Our models use the sum operator for each type of graph $\mathcal{G}^{(d)}$
inspired by the ``Device aggregation'' and ``Activity aggregation'' patterns, as well as use
the average operator for different types of graphs.

\subsection{Attention Mechanism}
Attention mechanisms have proven to be effective in many sequence-based~\cite{bahdanau2014neural} or
image-based tasks~\cite{desimone1995neural}. While we are
dealing with different types of devices, typically we are unknown of the importance of
the transformed information comes from different subgraphs $\mathcal{G}^{(d)}$. Instead of
simply averaging the information together in Eq.~\eqref{eq:estep},
we adaptively estimate the attention coefficients in the learning procedure for different types
of subgraphs. That is, we have:
\begin{align}
  H^{(t)} \leftarrow \sigma \bigg(X\cdot W +
  \sum_{d \in \mathcal{D}} \mathrm{softmax}(\alpha_d)\cdot\,\,\,\,\,A^{(d)}\cdot H^{(t-1)}\cdot V_d\bigg)
\end{align}
where $\mathrm{softmax}(\alpha_d) = \frac{\exp \alpha_d}{\sum_i \exp \alpha_i}$,
and $\alpha = [\alpha_1, ..., \alpha_{|\mathcal{D}|}]^\top \in \mathbb{R}^{|\mathcal{D}|}$
is a free parameter need to be estimated.

\section{Experiments}
In this section we show the experimental results of our approaches
deployed as a real system at Alipay. 

\subsection{Datasets}

\begin{table}
  \caption{Experimental Data Summary.}
  \label{tb:data}
  \begin{tabular}{cc}
    \toprule
    & Count\\
    \midrule
    \#Vertices & $8\times 10^6$  \\
    \#Edges & $10 \times 10^6$\\
    \#Labels in train & $17\times 10^5$ \\
    \#Labels in test & $2 \times 10^5$ \\
    \#Features & $374$ \\
  \bottomrule
\end{tabular}
\end{table}

We deploy our approach at Alipay\footnote{{https://en.wikipedia.org/wiki/Ant\_Financial}},
the world's leading mobile
payment platform serving more than 450 millions of users. Our system
targets on hundred thousands of newly registered accounts daily. For those accounts
already been used in a long term, it is much trivial to identify their risks because
we have already collected enough profiles for risk evaluations.
To predict newly registered accounts daily, everyday we build the graph
using all the active accounts and associated devices generated
from past 7 days. We further \emph{preprocess}
the data by deleting the accounts connected to devices shared with no other accounts,
i.e. isolated nodes. Such accounts are either in a very low risk of being malicious ones,
or useless in propagating informations through the topology. Thus we use
the rest accounts and assoicated devices as vertices in the \emph{preprocessed} data.

To show the effectiveness, in our experiments, we use a period of one month \emph{preprocessed} dataset at Alipay.
The rough statistics of the experimental dataset are summarized in Table~\ref{tb:data}.
We split the data into 4 consecutive weeks, namly, ``week 1'', ``week 2'',
``week 3'' and ``week 4''. For each week, we build the heterogeneous graph
using the vertices (accounts and devices) and associated
edges (activities) during that week.
All the partially labeled accounts come from the first 6 days, and we aim to
predict the accounts newly registered at the end of each week. We show the
results from consecutive 4 weeks for the purpose
of robustness. Due to the policy of information sensitivity at Alipay, we will not
reveal the ratio of malicious accounts and normal accounts
because those numbers are extremely sensitive.

To get the activity features $x_i$, we discretize
the activities in hours, i.e. $p = 7 \times 24 = 168$ slots, with the
value of each slot as the counts of $i$ having activities
in the time slot. In addition, we have 6 types of devices as discussed in section~\ref{sec:het},
as well as around 200 demographics features for each account,
thus results into $374=168+6+200$ dimensional features.


\subsection{Experimental Settings}
We describe our experimental settings as follows.

{\bfseries Evaluation}.
Alipay first identifies suspicious newly registered
accounts and observes those accounts in a long term.
Afterwards, Alipay is able to give ``ground truth'' labels to those accounts
with the benefit of hindsight.
In the following sections, we will report the
F-1\footnote{\url{https://en.wikipedia.org/wiki/F1_score}} and
AUC\footnote{\url{http://fastml.com/what-you-wanted-to-know-about-auc/}} measure,
and evaluate the precision and recall curve on such ``ground
truth'' labels.

The reason we care about precision and recall curve is that it is
required that the system should be able to detect malicious
with high confidence at least at the top of scored suspicious accounts,
so that the system will not interrupt and disturb most of normal users. This is quite important
for an Internet business company providing financial services.
On the other hand, we would like to avoid huge capital loss as possible
at the same time. The precision and recall curves can tell under
which threshold, our detection system could well-balance the service experiences
and cover ratio of malicious accounts. Note that this is quite different
from the threshold set as $0.5$ in academia.

{\bfseries Comparison Methods}.
We compare our methods with four baseline methods.
\begin{itemize}
\item Connected Subgraph, which is discussed in section~\ref{sec:mot}. This
  approach is similar to the approach introduced in
~\cite{zhao2009botgraph}. The method first builds an account-account graph,
and we define the weight of each edge as the inner product of
two accounts $x_i$ and $x_{i'}$. 
The measure of such affinity can help us split out
normal accounts in a giant connected subgraph, to further balance
the trade-off between precision and recall. Finally, we treat the
corresponding component size as the score of each account.
\item GBDT+Graph, which is a machine learning-based method, that
  we first calculate the statistic properties of the account-account graph,
  e.g. the connected subgraph component size, the in-degree, out-degree of
  each account, along with features of each account, we feed those
  features to a very competitive classifier Gradient Boosting Decision Tree~\cite{chen2016xgboost} (GBDT)
  which is widely used in industry.
\item GBDT+Node2Vec~\cite{grover2016node2vec}, which is a type of random walk-based node
  embedding methods described in section~\ref{sec:node2vec}.
  The unsupervised method first learn representations of each node in our
  device-account graph with the purpose of preserving the topology of the graph.
  After that, we feed the learned embeddings along with original features
  to a GBDT classifier. We treat all devices as the same type because this method
  cannot deal with heterogeneous graph trivially.
\item GCN~\cite{kipf2016semi}, which is one of the classic graph convolutional network
  based approach, that it aggregates the neighborhood as per Eq.~\eqref{eq:convolution}.
\end{itemize}

For graph convolutional network-style methods including our methods, we set
embedding size as 16 with a depth of the convolution layers as 5, unless otherwise stated.
For GBDT, we use 100 trees with learning rate as 0.1. For Node2Vec~\cite{grover2016node2vec}, we repeatedly sample
100 paths for each node, with the length of each path as 50.

\begin{table}
  \caption{F-1 Score}
  \label{tb:f1}
  \begin{tabular}{ccccc}
    \toprule
    &week 1&week 2&week 3&week 4\\
    \midrule
     Connected Subgraphs & 0.5033  & 0.5567 & 0.58 & 0.5421 \\
     GBDT+Graph & 0.7423 & 0.7598 & 0.7693 & 0.6639 \\
     GBDT+Node2Vec & 0.741 & 0.7571 & 0.769 & 0.6626 \\
     GCN & 0.7729 & 0.7757 & 0.7957 & 0.6919 \\
     GEM (Ours) & 0.7992 & 0.8066 & 0.8191 & 0.718\\
     GEM-attention (Ours) & \bf{0.8165} & \bf{0.8133} & \bf{0.8244} & \bf{0.7344}\\
  \bottomrule
\end{tabular}
\end{table}

\begin{table}
  \caption{AUC}
  \label{tb:auc}
  \begin{tabular}{ccccc}
    \toprule
    &week 1&week 2&week 3&week 4\\
    \midrule
     Connected Subgraphs & 0.6689  & 0.6692 & 0.665 & 0.6938 \\
     GBDT+Graph & 0.8878 & 0.8835 & 0.8707 & 0.8778 \\
     GBDT+Node2Vec & 0.8884 & 0.883 & 0.8711 & 0.8773 \\
     GCN & 0.8995 & 0.8932 & 0.8922 & 0.881 \\
     GEM (Ours) & 0.9159 & 0.9238 & 0.9193 & 0.9082\\
     GEM-attention (Ours) & \bf{0.9364} & \bf{0.9293} & \bf{0.9259} & \bf{0.9155}\\
  \bottomrule
\end{tabular}
\end{table}

\begin{figure*}
\centering
\subfigure { \includegraphics[width=0.46\textwidth, height=0.31\textwidth]{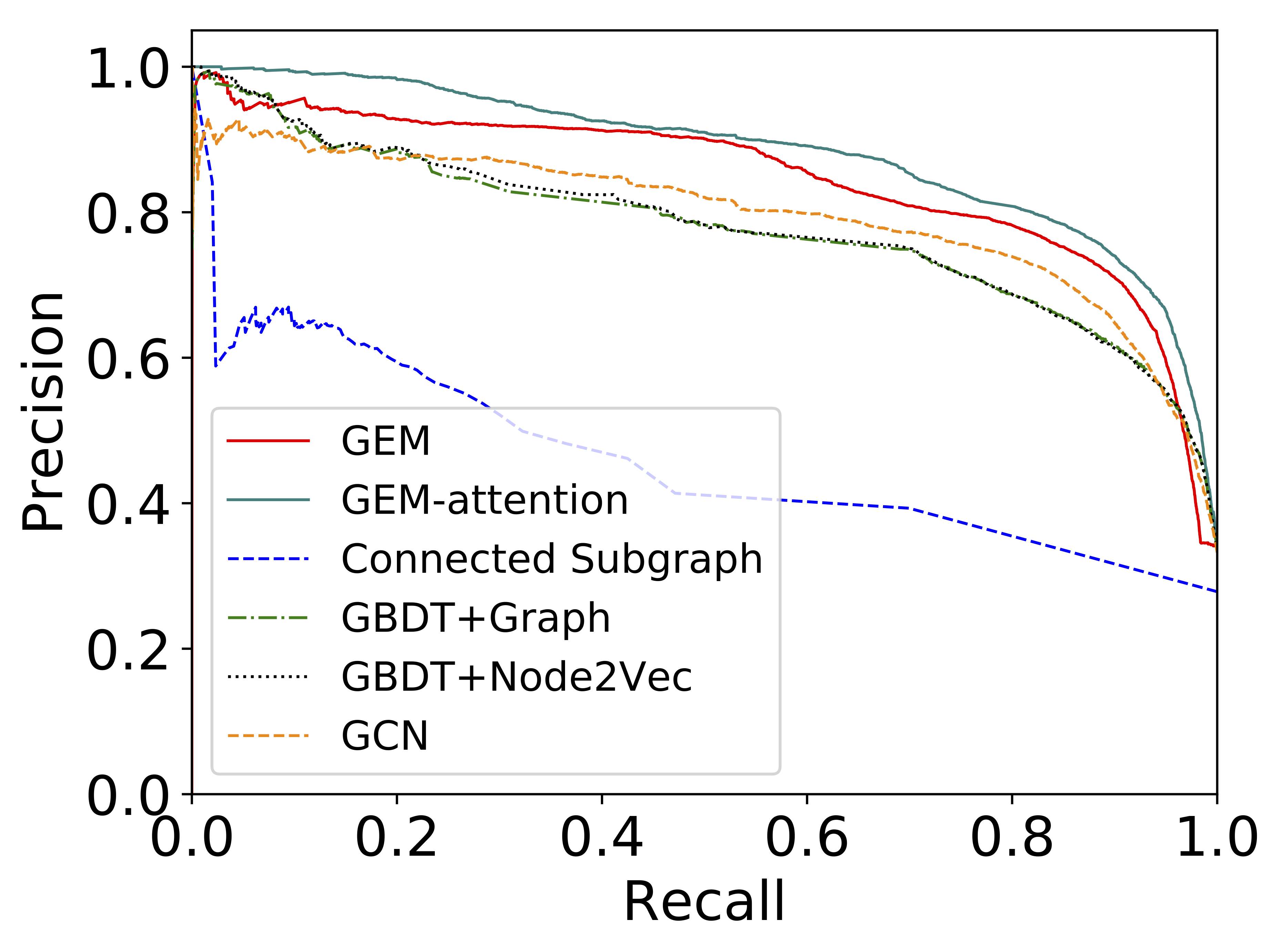}}
\subfigure { \includegraphics[width=0.46\textwidth, height=0.31\textwidth]{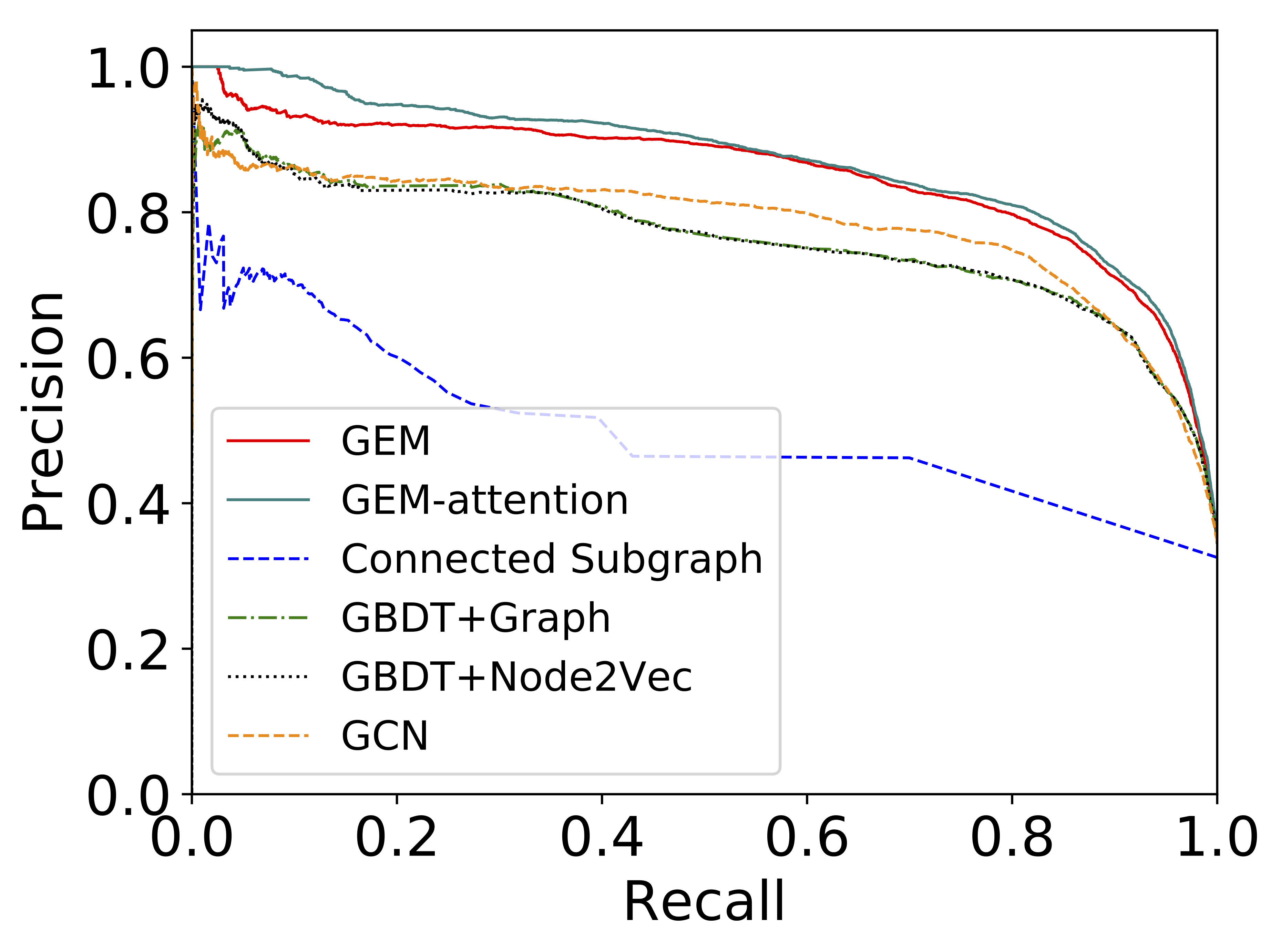}}
\subfigure { \includegraphics[width=0.46\textwidth, height=0.31\textwidth]{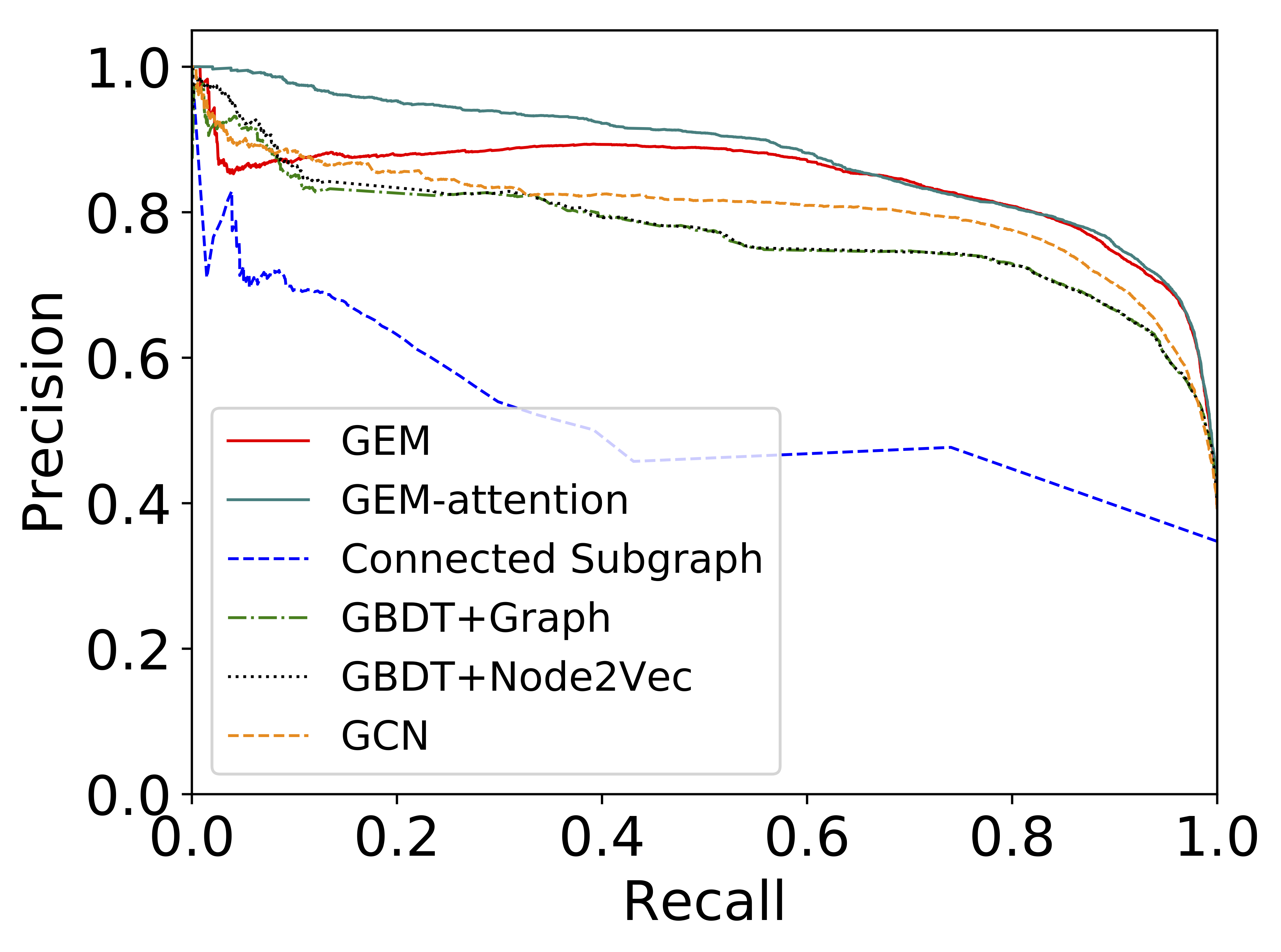}}
\subfigure { \includegraphics[width=0.46\textwidth, height=0.31\textwidth]{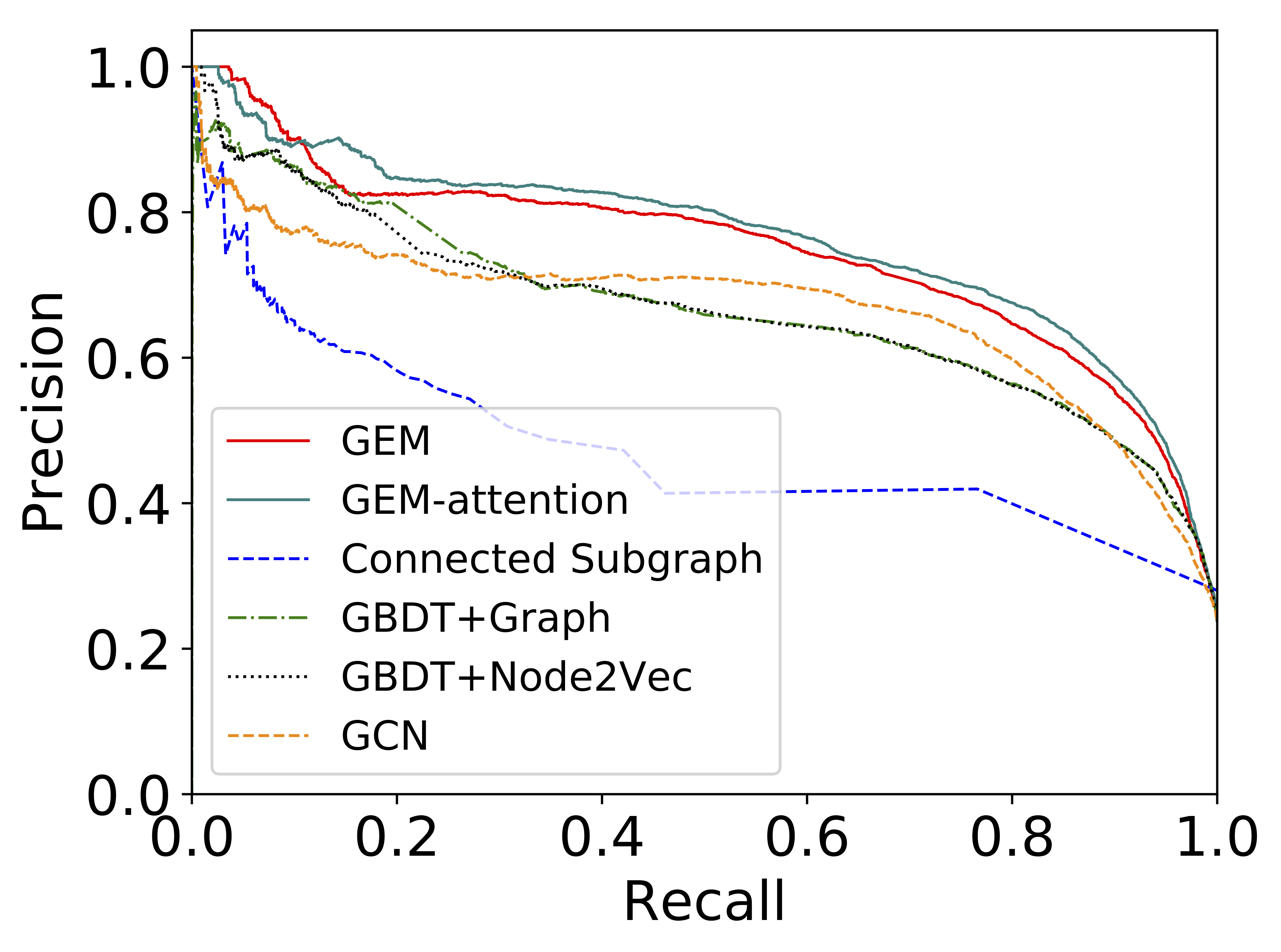}}
\caption{Precision-Recall Curves of GEM, GEM-attention, Connected Subgraph,
  GBDT+Graph, GBDT+Node2Vec and GCN on test data: week 1 (\emph{first}), 
  week 2 (\emph{second}), week 3 (\emph{third}), week 4 (\emph{fourth}) .}
\label{fig:pr}
\end{figure*}

\subsection{Results}
\subsubsection{Basic Measures}
We first report the results of all the methods in terms of standard
classification measures, i.e. F-1 scores
and AUC in Table~\ref{tb:f1} and Table~\ref{tb:auc}.

As can be seen, even though the connected subgraph component method is quite
intuitive, they are not doing well on this classification problem. The reason is
apparrent that large amount of benign accounts interwined with malicious accounts
in the device-account graph due to noisy data in practice. There are malicious
accouts exist both in large and small connected subgraphs.

The result of GBDT+Graph method is quite
similar compared with GBDT+Node2Vec. This might be essentially Node2Vec aims to
learn the properties of the graph which is similar to our features extracted
in GBDT+Graph.

GCN works better than GBDT+Graph and GBDT+Node2Vec. The reason might be:
GCN directly learns node embeddings using the responses of labels and activity features,
while the embeddings from Node2Vec or the graph statistics are not optimized for
the labels.

Our method GEM consistently outperforms GCN. The reason is two-folds:
(1) GEM deals with heterogeneous types of devices compared with GCN that can only deal
with homogeneous graphs that GCN can not discern the different types of nodes in graph;
(2) GEM uses aggregator operator for each type of nodes instead of normalized operator~\cite{kipf2016semi}
so that it can well model the underlying aggregation patterns as we discussed in section~\ref{sec:example}.

Finally, we find our GEM-attention with attention mechanism that adaptively assigns different
attention coefficients to each type of device network performs the best.
This is due to the reason that instead of normalizing each type of devices as the same of
importance, we should learn their importances from our data because (1) the different types
of devices might be noisy in different degrees, for example, IP addresses might be easily
confused while UMID could be more unique and accurate; (2) The certain device data could be potentially missing.

\subsubsection{Precision-Recall Curves}
We report the Precision-Recall curve of all the methods in Figure~\ref{fig:pr}.
As we can see, our proposed method GEM significantly outperforms
the comparison methods in terms of the area beneath the Precision-Recall curve.

One of the largest connected subgraph consists of a total of 1538 accounts
aggregating together in our experimental dataset.
The connected subgraph method can precisely identify most of accounts in the largest
connected subgraphs as malicious accounts due to the strong signal.
This leads to high precision at the very begining of the curve.
After that, the precision of the connected subgraph method drops
quickly. That is, it is extremely hard for such methods to
retain consistent high precision/recall curves when the size of
identified connected subgraphs
tends to be small.

Our methods work similar or even better at the very begining of
the curve compared with the comparison methods. More importantly,
our methods can accurately detect much more malicious accounts (high recall)
with still relative high precision, which is quite promising.

\subsubsection{Model Complexity}
In this section, we study the model complexity includes embedding sizes,
the depth of hidden convolution layers, and their impact on our task.

{\bf Varying Embedding Sizes.}
We vary embedding sizes from 8, 16, 32, 64 to 128. With larger embedding sizes,
we need to add slightly stronger regularizers on our models. With appropriate
regularizers, we do not find significant differences in terms of F-1 score.

\begin{figure}
\centering
\subfigure { \includegraphics[width=0.3\textwidth, height=0.25\textwidth]{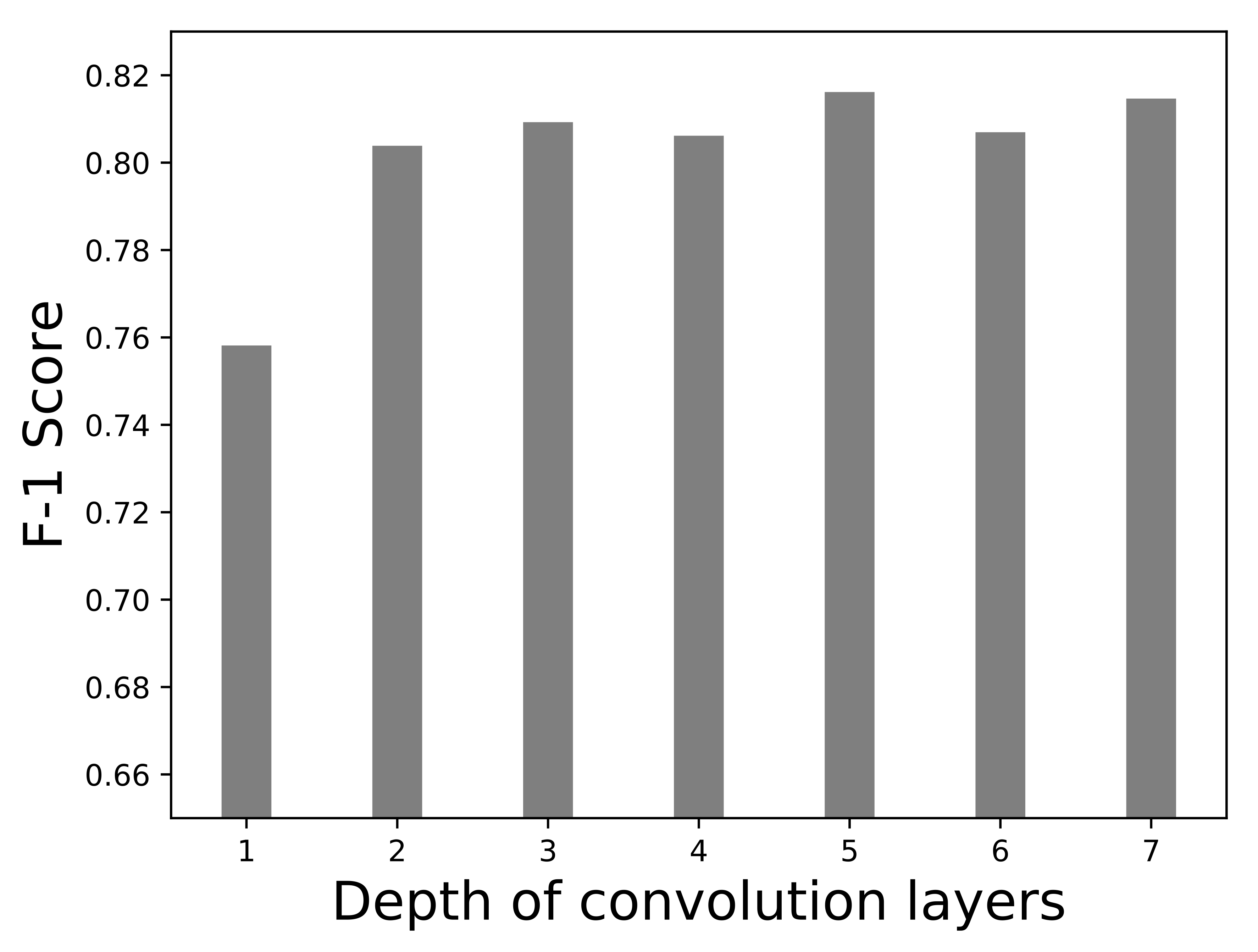}}
\caption{F-1 Score while varying the depths of hidden layers.}
\label{fig:depth}
\end{figure}

{\bf Varying the Depths of Layers.}
Indeed, the depth of our hidden convolution layers influences the F-1 scores
quite a lot. With deeper hidden layers, our model tends to aggregate transformed
information from a neighborhood to a greater extent. We show the F-1 scores with
varying depths of hidden layers in Figure~\ref{fig:depth}.

The F-1 score with a depth of 1 hidden layer does not work well because of the
heterogeneous graphs we have. Our model needs to ``exchange'' information among
accounts via devices, that requires at least two hops of neighbors to look at.

\subsubsection{Attention Coefficients}\label{sec:attention}
In this section, we study the contributions of each type of
devices in identifying the malicious accounts by illustrating
the estimated attention coefficients using the dataset ``week 1''.
We show those assigned attention coefficients in table~\ref{tb:assigned}.
The results show that different types of nodes in a heterogeneous graph
could have different impacts on the identification of malicious accounts.

We illustrate one of connected subgraphs with the thicknesses of
edges as the corresponding attention coefficients in Figure~\ref{fig:visual}.
%
\begin{table}
  \caption{Assigned attention coefficients estimated using dataset ``week 1''.}
  \label{tb:assigned}
  \begin{tabular}{cc}
    \toprule
    Devices & Attention coefficients\\
    \midrule
    UMID & 0.4412 \\
    Phone Number & 0.2952\\
    MAC& 0.13 \\
    APDID & 0.1068 \\
    IMSI & 0.0142 \\
    TID &0.0125 \\
  \bottomrule
\end{tabular}
\end{table}

\subsubsection{Online Results}
In practice, everyday we treat top ten thousand scored newly registered accounts identified by our
approach as accounts at risk. Under this strategy, the precision evaluated by the security department from Alipay
is over 98\% after a long time observation. Compared with a former
deployed rule-based approach, our GEM can cover 10\% more accounts.
Thus, we are able to capture more high risk accounts while maintaining
very competitive precision.

\section{Conclusion}
In this paper, we show our experiences on designing novel graph
neural networks to detect ten thousands malicious accounts daily at Alipay.
In particular, we summarize two fundamental weaknesses of attackers, namely
``Device aggregation'' and ``Activity aggregation'', and
naturally present a neural network approach based on heterogeneous account-device graphs.
This is the first work that graph neural network approach has ever been applied
to fraud detection problems.
Our methods achieve promising precision-recall curves compared
with competitive methods.
Furthermore, we discuss the ideas of re-formulating the intuitive connected subgraph approach to
our graph neural network approach.
In future, we are interested in
building a real-time malicious account detection system
based on dynamic graphs instead of the proposed daily detection
system.


\end{document}